\documentclass[11pt,a4paper]{article}
\usepackage[hyperref]{emnlp-ijcnlp-2019}
\usepackage{times}
\usepackage{latexsym}
\usepackage{adjustbox}
\usepackage{amsmath,amssymb,amsthm}
\usepackage{enumitem}
\usepackage{graphicx}  
\usepackage{url}
\usepackage{caption}
\usepackage{mathtools}
\usepackage{footnote}
\usepackage[ruled,vlined,linesnumbered]{algorithm2e}
\makesavenoteenv{table}

\aclfinalcopy 

\title{Leveraging Frequent Query Substructures to Generate Formal Queries \\ for Complex Question Answering}

\author{Jiwei Ding \quad Wei Hu\thanks{\ \ Corresponding authors} \quad Qixin Xu \quad Yuzhong Qu$^\ast$ \\
  State Key Laboratory for Novel Software Technology, Nanjing University, China \\
  \texttt{\{jwding, qxxu\}nju@outlook.com, \{whu, yzqu\}@nju.edu.cn} }

\date{}

\begin{document}
\maketitle
\begin{abstract}
Formal query generation aims to generate correct executable queries for question answering over knowledge bases (KBs), given entity and relation linking results. Current approaches build universal paraphrasing or ranking models for the whole questions, which are likely to fail in generating queries for complex, long-tail questions. In this paper, we propose SubQG, a new query generation approach based on frequent query substructures, which  helps rank the existing (but nonsignificant) query structures or build new query structures. Our experiments on two benchmark datasets show that our approach significantly outperforms the existing ones, especially for complex questions. Also, it achieves promising performance with limited training data and noisy entity/relation linking results.
\end{abstract}

\section{Introduction}
\begin{figure}[t]
\centering
\includegraphics[width=\columnwidth]{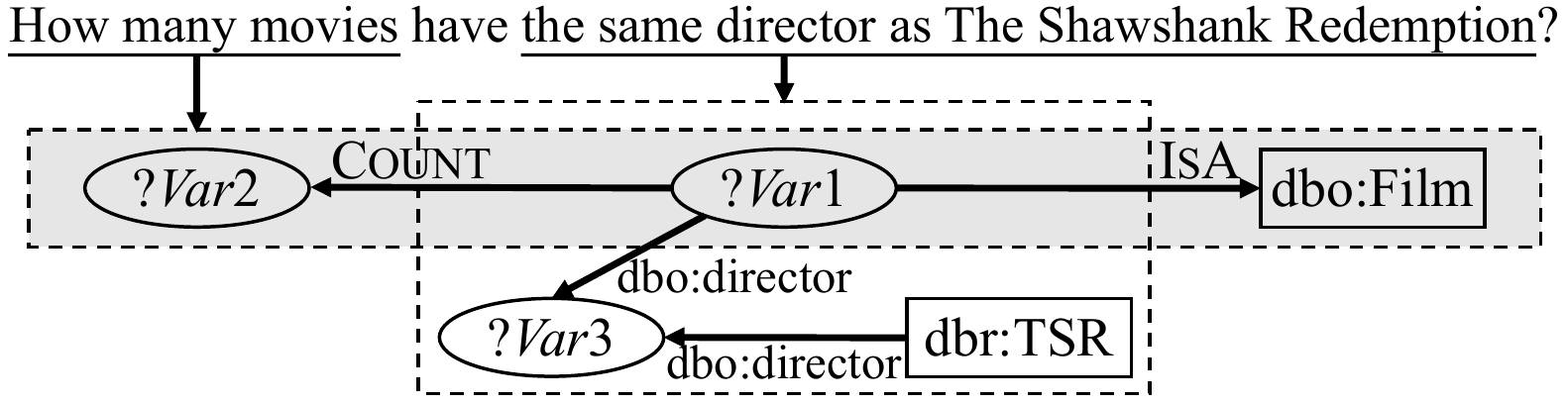}
\caption{An example for complex question and query}
\label{fig:intro}
\end{figure}

Knowledge-based question answering (KBQA) aims to answer natural language questions over knowledge bases (KBs) such as DBpedia and Freebase. Formal query generation is an important component in many KBQA systems \cite{ConstraintQG,KBQA,CQAEMNLP}, especially for answering complex questions. Given entity and relation linking results, formal query generation aims to generate correct executable queries, e.g., SPARQL queries, for the input natural language questions. An example question and its formal query are shown in Figure \ref{fig:intro}. Generally speaking, formal query generation is expected to include but not be limited to have the capabilities of (\romannumeral1) recognizing and paraphrasing different kinds of constraints, including triple-level constraints (e.g., \emph{``movies"} corresponds to a typing constraint for the target variable) and higher level constraints (e.g., subgraphs). For instance, \emph{``the same ... as"} represents a complex structure shown in the middle of Figure \ref{fig:intro}; (\romannumeral2) recognizing and paraphrasing aggregations (e.g., ``how many" corresponds to \textsc{Count}); and (\romannumeral3) organizing all the above to generate an executable query \cite{QueryBuilding, SQG}.

There are mainly two kinds of query generation approaches for complex questions. (\romannumeral1) Template-based approaches choose a pre-collected template for query generation \cite{KBQA,TemplateQAWWW17}. Such approaches highly rely on the coverage of templates, and perform unstably when some complex templates have very few natural language questions as training data.
(\romannumeral2) Approaches based on semantic parsing and neural networks learn entire representations for questions with different query structures, by using a neural network following the encode-and-compare framework \cite{CQAEMNLP,SQG}. They may suffer from the lack of training data, especially for long-tail questions with rarely appeared structures. Furthermore, both above approaches cannot handle questions with unseen query structures, since they cannot generate new query structures.

To cope with the above limitations, we propose a new query generation approach based on the following observation: the query structure for a complex question may rarely appear, but it usually contains some substructures that frequently appeared in other questions. For example, the query structure for the question in Figure \ref{fig:intro} appears rarely, however, both \emph{``how many movies"} and \emph{``the same ... as"} are common expressions, which correspond to the two query substructures in dashed boxes. To collect such frequently appeared substructures, we automatically decompose query structures in the training data. Instead of directly modeling the query structure for the given question as a whole, we employ multiple neural networks to predict query substructures contained in the question, each of which delivers a part of the query intention. Then, we select an existing query structure for the input question by using a combinational ranking function. Also, in some cases, no existing query structure is appropriate for the input question. To cope with this issue, we merge query substructures to build new query structures.
The contributions of this paper are summarized below:
\begin{itemize}\setlength{\itemsep}{0pt}
  \item We formalize the notion of query structures and define the substructure relationship between query structures.
  \item We propose a novel approach for formal query generation, which firstly leverages multiple neural networks to predict query substructures contained in the given question, and then ranks existing query structures by using a combinational function.
  \item We merge query substructures to build new query structures, which handles questions with unseen query structures.
  \item We perform extensive experiments on two KBQA datasets, and show that SubQG significantly outperforms the existing approaches. Furthermore, SubQG achieves a promising performance with limited training data and noisy entity/relation linking results.
\end{itemize}
\section{Preliminaries}
\label{sect:pre}
An entity is typically denoted by a URI and described with a set of properties and values. A fact is an $\langle entity, property, value\rangle$ triple, where the value can be either a literal or another entity. A KB is a pair $\mathcal{K}=(\mathcal{E},\mathcal{F})$, where $\mathcal{E}$ denotes the set of entities and $\mathcal{F}$ denotes the set of facts.

A \emph{formal query} (or simply called \emph{query}) is the structured representation of a natural language question executable on a given KB. Formally, a query is a pair $\mathcal{Q}=(\mathcal{V},\mathcal{T})$, where $\mathcal{V}$ denotes the set of vertices, and $\mathcal{T}$ denotes the set of labeled edges. A vertex can be either a variable, an entity or a literal, and the label of an edge can be either a built-in property or a user-defined one. For simplicity, the set of all edge labels are denoted by $\mathcal{L}_e(\mathcal{Q})$. In this paper, the built-in properties include \textsc{Count}, \textsc{Avg}, \textsc{Max}, \textsc{Min}, \textsc{MaxAtN}, \textsc{MinAtN} and \textsc{IsA} (\textsc{rdf:type}), where the former four are used to connect two variables. For example, $\langle?Var1,\textsc{Count},?Var2\rangle$ represents that $?Var2$ is the counting result of $?Var1$. \textsc{MaxAtN} and \textsc{MinAtN} take the meaning of \textsc{Order By} in SPARQL \cite{ConstraintQG}. For instance, $\langle?Var1,\textsc{MaxAtN},2\rangle$ means \textsc{Order By} \textsc{Desc}$(?Var1)$ \textsc{Limit 1 Offset 1}.

To classify various queries with similar query intentions and narrow the search space for query generation, we introduce the notion of \emph{query structures}. A query structure is a set of structurally-equivalent queries. Let $\mathcal{Q}_a=(\mathcal{V}_a,\mathcal{T}_a)$ and $\mathcal{Q}_b=(\mathcal{V}_b,\mathcal{T}_b)$ denote two queries. $\mathcal{Q}_a$ is structurally-equivalent to $\mathcal{Q}_b$, denoted by $\mathcal{Q}_a\cong\mathcal{Q}_b$, if and only if there exist two bijections $f:\mathcal{V}_a\rightarrow\mathcal{V}_b$ and $g:\mathcal{L}_e(\mathcal{Q}_a)\rightarrow\mathcal{L}_e(\mathcal{Q}_b)$ such that:
\begin{enumerate}[label=(\roman*),itemsep=0pt]
\item $\forall v\in\mathcal{V}_a$, $v$ is a variable $\Leftrightarrow f(v)$ is a variable;
\item $\forall r\in\mathcal{L}_e(\mathcal{Q}_a)$, $r$ is a user-defined property $\Leftrightarrow g(r)$ is a user-defined property; if $r$ is a built-in property, $g(r)=r$;
\item $\forall v\forall r\forall v'  \langle v,r,v'\rangle\in\mathcal{T}_a$ $\Leftrightarrow$ $\langle f(v),g(r),f(v')\rangle$ $\in\mathcal{T}_b$.
\end{enumerate}

\begin{figure}
\centering
\includegraphics[width=\columnwidth]{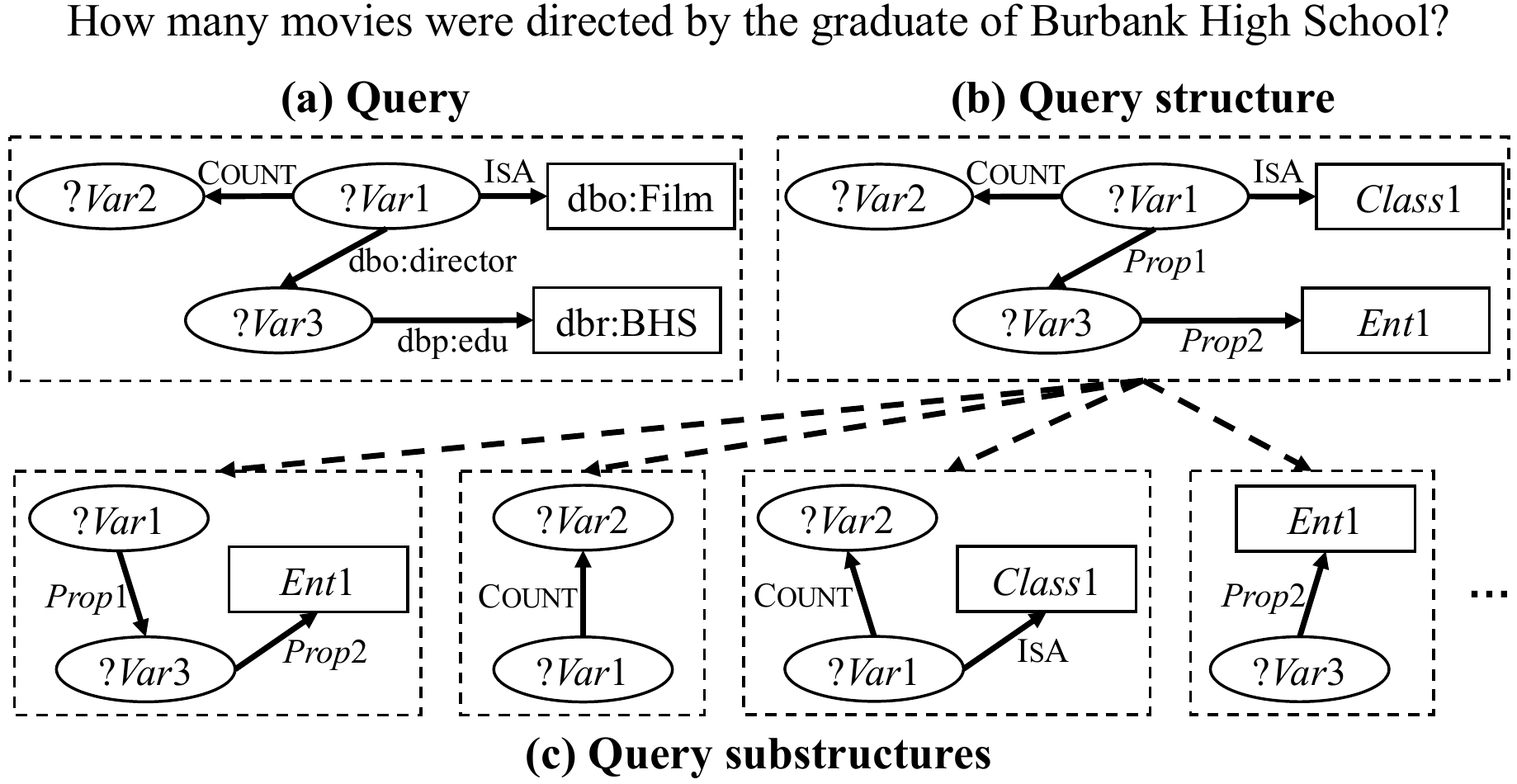}
\caption{Illustration of a query, a query structure and query substructures}
\label{fig:query}
\end{figure}

The query structure for $\mathcal{Q}_a$ is denoted by $\mathcal{S}_a=[\mathcal{Q}_a]$, which contains all the queries structurally-equivalent to $\mathcal{Q}_a$. For graphical illustration, we represent a query structure by a representative query among the structurally-equivalent ones and replace entities and literals with different kinds of placeholders. An example of query and query structure is shown in the upper half of Figure~\ref{fig:query}.

For many simple questions, two query structures, i.e., $(\{?Var1,Ent1\}, \{\langle ?Var1,Prop1,$ $Ent1\rangle\})$ and $(\{?Var1,Ent1\},\{\langle Ent1,Prop1,$ $?Var1\rangle\})$, are sufficient. However, for complex questions, a diversity of query structures exist and some of them share a set of frequently-appeared substructures, each of which delivers a part of the query intention. We give the definition of \emph{query substructures} as follows.

Let $\mathcal{S}_a=[\mathcal{Q}_a]$ and $\mathcal{S}_b=[\mathcal{Q}_b]$ denote two query structures. $\mathcal{S}_a$ is a query substructure of $\mathcal{S}_b$, denoted by $\mathcal{S}_a\preceq\mathcal{S}_b$, if and only if $\mathcal{Q}_b$ has a subgraph $\mathcal{Q}_c$ such that $\mathcal{Q}_a\cong\mathcal{Q}_c$. Furthermore, if $\mathcal{S}_a=[\mathcal{Q}_a]\preceq\mathcal{S}_b=[\mathcal{Q}_b]$, we say that $\mathcal{Q}_b$ has $\mathcal{S}_a$, and $\mathcal{S}_a$ is contained in $\mathcal{Q}_b$.

For example, although the query structures for the two questions in Figures~\ref{fig:intro} and \ref{fig:query} are different, they share the same query substructure $(\{?Var1,$ $?Var2,Class1\},\{\langle ?Var1,\textsc{Count},?Var2\rangle,$ $\langle ?Var1, \textsc{IsA}, Class1\rangle\})$, which corresponds to the phrase ``how many movies". Note that, a query substructure can be the query structure of another question.

The goal of this paper is to leverage a set of frequent query (sub-)structures to generate formal queries for answering complex questions.

\section{The Proposed Approach}
In this section, we present our approach, SubQG, for query generation. We first introduce the framework and general steps with a running example (Section \ref{sec:framework}), and then describe some important steps in detail in the following subsections.

\subsection{Framework}
\label{sec:framework}
\begin{figure}[t]
\centering
\includegraphics[width=\columnwidth]{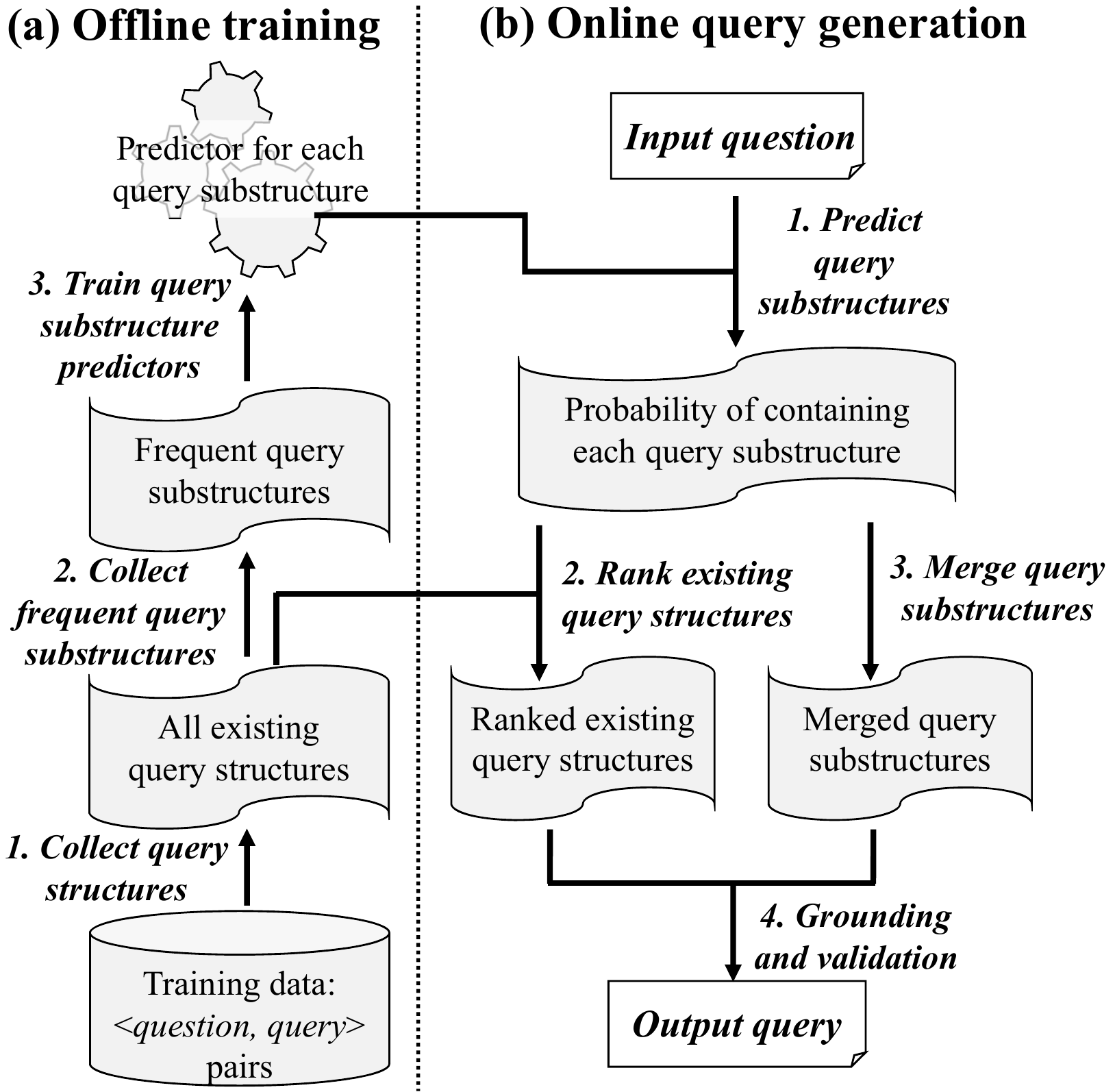}
\caption{Framework of the proposed approach}
\label{fig:framework}
\end{figure}
Figure \ref{fig:framework} depicts the framework of SubQG, which contains an offline training process and an online query generation process.

\textbf{Offline.} The offline process takes as input a set of training data in form of $\langle question, query \rangle $ pairs, and mainly contains three steps:\\
\textbf{1. Collect query structures.} For questions in the training data, we first discover the structurally-equivalent queries, and then extract the set of all query structures, denoted by $\mathbf{TS}$.\\
\textbf{2. Collect frequent query substructures.} We decompose each query structure $\mathcal{S}_i=(\mathcal{V}_i, \mathcal{T}_i) \in \textbf{TS}$ to get the set for all query substructures. Let $\mathcal{T}_j$ be a non-empty subset of $\mathcal{T}_i$, and $\mathcal{V}_{\mathcal{T}_j}$ be the set of vertices used in $\mathcal{T}_j$. $\mathcal{S}_j=(\mathcal{V}_{\mathcal{T}_j}, \mathcal{T}_j)$ should be a query substructure of $\mathcal{S}_i$ according to the definition. So, we can generate all query substructures of $\mathcal{S}_i$ from each subset of $\mathcal{T}_i$. Disconnected query substructures would be ignored since they express discontinuous meanings and should be split into smaller query substructures. If more than $\gamma$ queries in training data have substructure $\mathcal{S}_j$, we consider $\mathcal{S}_j$ as a frequent query substructure. The set for all frequent query substructures is denoted by $\mathbf{FS^*}$.\\
\textbf{3. Train query substructure predictors.} We train a neural network for each query substructure $\mathcal{S}^*_i \in \mathbf{FS^*}$, to predict the probability that $\mathcal{Q}^y$ has $\mathcal{S}^*_i$ (i.e., $\mathcal{S}^*_i \preceq [\mathcal{Q}^y]$) for input question $y$, where $\mathcal{Q}^y$ denotes the formal query for $y$. Details for this step are described in Section \ref{sec:subPredict}.

\begin{figure}[t]
\centering
\includegraphics[width=\columnwidth]{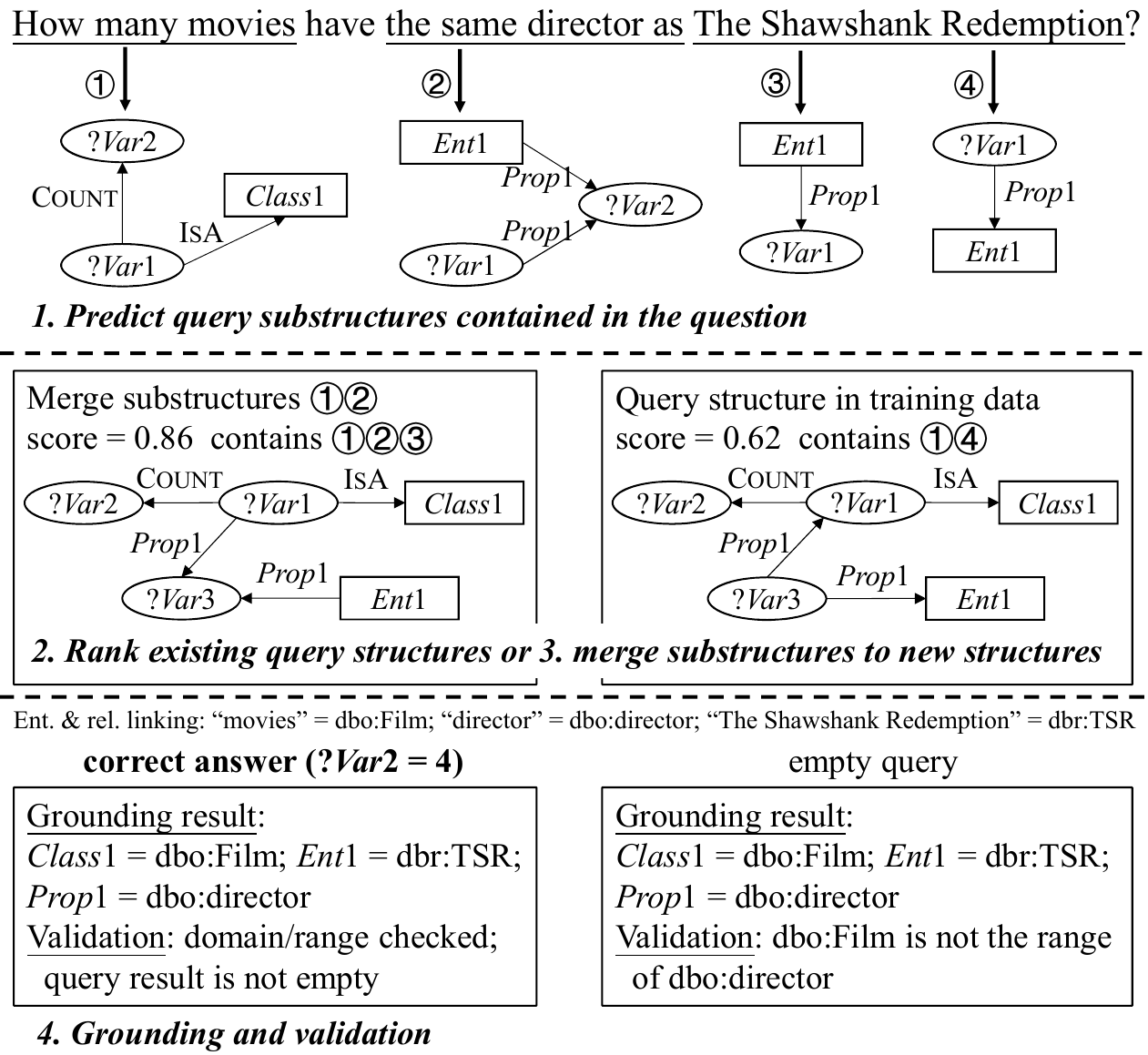}
\caption{An example for online query generation}
\label{fig:online}
\end{figure}
\textbf{Online.} The online query generation process takes as input a natural language question $y$, and mainly contains four steps:\\
\textbf{1. Predict query substructures.} We first predict the probability that $\mathcal{S}^*_i \preceq [\mathcal{Q}^y]$ for each $\mathcal{S}^*_i \in \mathbf{FS^*}$, using the query substructure predictors trained in the offline step. An example question and four query substructures with highest prediction probabilities are shown in the top of Figure \ref{fig:online}.\\
\textbf{2. Rank existing query structures.} To find an appropriate query structure for the input question, we rank existing query structures ($\mathcal{S}_i\in \mathbf{TS}$) by using a scoring function, see Section \ref{sec:structureRank}.\\
\textbf{3. Merge query substructures.} Consider the fact that the target query structure $[\mathcal{Q}^y]$ may not appear in $\mathbf{TS}$ (i.e., there is no query in the training data that is structurally-equivalent to $\mathcal{Q}^y$), we design a method (described in Section \ref{sec:merge}) to merge question-contained query substructures for building new query structures. The merged results are ranked using the same function as existing query structures.
Several query structures (including the merged results and the existing query structures) for the example question are shown in the middle of Figure \ref{fig:online}.\\
\textbf{4. Grounding and validation.} We leverage the query structure ranking result, alongside with the entity/relation linking result from some existing black box systems \cite{EARL} to generate executable formal query for the input question. For each query structure, we try all possible combinations of the linking results according to the descending order of the overall linking score, and perform validation including grammar check, domain/range check and empty query check. The first non-empty query passing all validations is considered as the output for SubQG. The grounding and validation results for the example question are shown in the bottom of Figure \ref{fig:online}.

\subsection{Query Substructure Prediction}
\label{sec:subPredict}

In this step, we employ an attention based Bi-LSTM network \cite{Attention} to predict $\textrm{Pr}[\mathcal{S}^*_i\,|\,y]$ for each frequent query substructure $\mathcal{S}^*_i\in \mathbf{FS^*}$, where $\textrm{Pr}[\mathcal{S}^*_i\,|\,y]$ represents the probability of $\mathcal{S}^*_i \preceq [\mathcal{Q}^y]$. There are mainly three reasons that we use a predictor for each query substructure instead of a multi-tag predictor for all query substructures:
(\romannumeral1) a query substructure usually expresses part of the meaning of input question. Different query substructures may focus on different words or phrases, thus, each predictor should have its own attention matrix;
(\romannumeral2) multi-tag predictor may have a lower accuracy since each tag has unbalanced training data;
(\romannumeral3) single pre-trained query substructure predictor from one dataset can be directly reused on another without adjusting the network structure, however, the multi-tag predictor need to adjust the size of the output layer and retrain when the set of frequent query substructures changes.

\begin{figure}[t]
\centering
\includegraphics[width=\columnwidth]{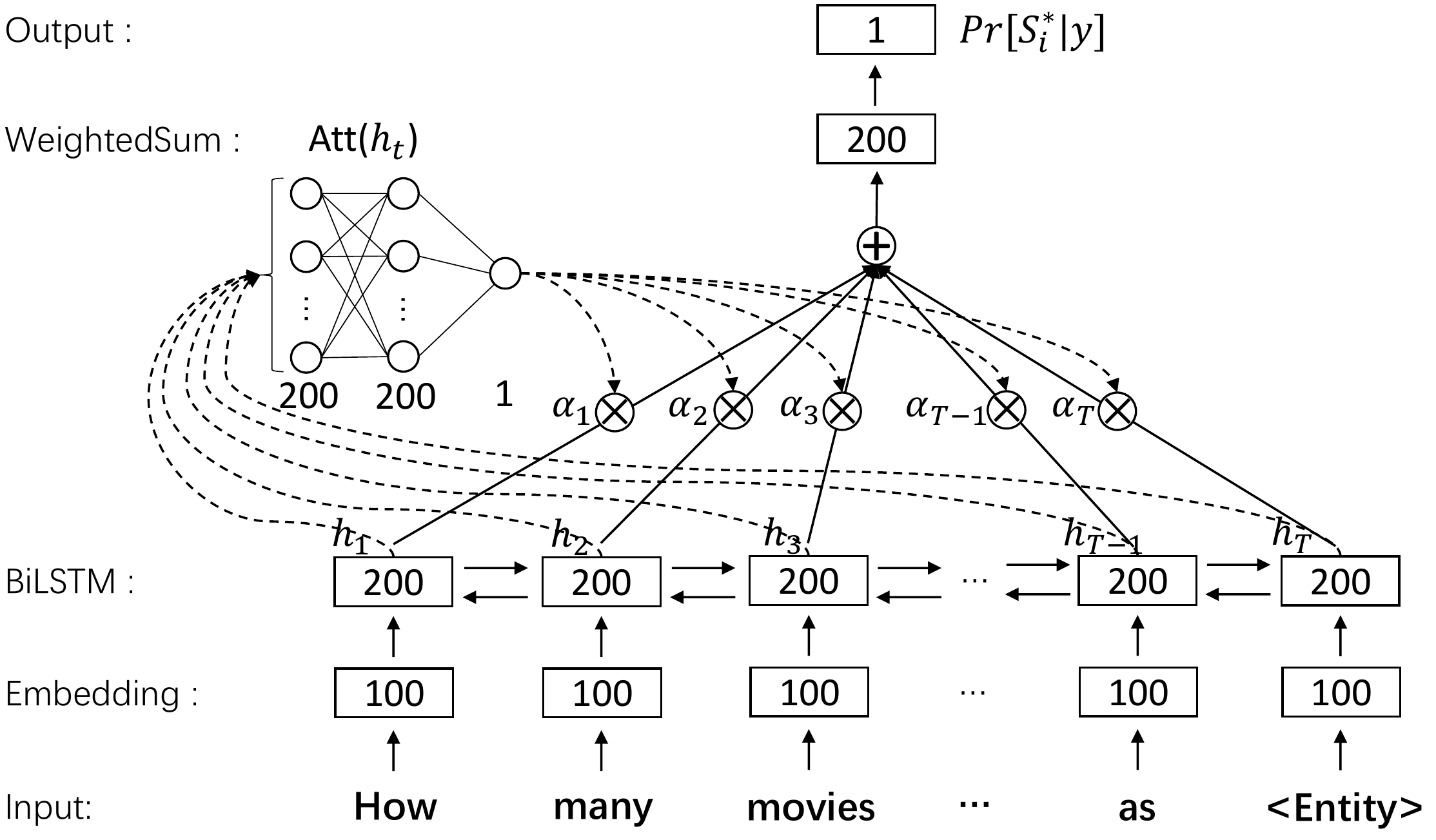}
\caption{Attention-based BiLSTM network}
\label{fig:network}
\end{figure}

The structure of the network is shown in Figure~\ref{fig:network}. Before the input question is fed into the network, we replace all entity mentions with \emph{ $\langle$Entity$\rangle$ } using EARL \cite{EARL}, to enhance the generalization ability. Given the question word sequence \{$w_1,...,w_T$\}, we first use a word embedding matrix to convert the original sequence into word vectors \{$\mathbf{e}_1,...,\mathbf{e}_T$\}, followed by a BiLSTM network to generate the context-sensitive representation \{$\mathbf{h}_1,...,\mathbf{h}_T$\} for each word, where
\begin{equation}
  \mathbf{h}_t = [\textsc{Lstm}(\mathbf{e}_{t},\overrightarrow{\mathbf{h}}_{t-1});\textsc{Lstm}(\mathbf{e}_{t},\overleftarrow{\mathbf{h}}_{t+1})] .
\end{equation}

Then, the attention mechanism takes each $\mathbf{h}_t$ as input, and calculates a weight $\alpha_t$ for each $\mathbf{h}_t$, which is formulated as follows:
\begin{align}
  \alpha_t &= \frac{e^{\textrm{Att}(\mathbf{h}_t)}}{\sum^T_{k=1}e^{\textrm{Att}(\mathbf{h}_k)}} ,\\
  \textrm{Att}(\mathbf{h}_t) &= \mathbf{v}_{att}^T\tanh(\mathbf{W}_{att}\mathbf{h}_t+\mathbf{b}_{att}),
\end{align}
where $\mathbf{W}_{att}\in\mathbb{R}^{|\mathbf{h}_t|\times|\mathbf{h}_t|}$, $\mathbf{b}_{att}\in\mathbb{R}^{|\mathbf{h}_t|}$ and $\mathbf{v}_{att}\in\mathbb{R}^{|\mathbf{h}_t|}$. Next, we get the representation for the whole question $\mathbf{q}^c$ as the weighted sum of $\mathbf{h}_t$:
\begin{equation}
  \mathbf{q}^c =  \sum^T_{t=1}\alpha_t \mathbf{h}_t .
\end{equation}

The output of the network is a probability
\begin{equation}
  \textrm{Pr}[\mathcal{S}^*_i\,|\,y] =  \sigma(\mathbf{v}_{out}^T\mathbf{q}^c+b_{out}) ,
\end{equation}
where $\mathbf{v}_{out}\in\mathbb{R}^{|\mathbf{q}^c|}$ and $b_{out}\in\mathbb{R}$. 

The loss function minimized during training is the binary cross-entropy:
\begin{equation}
\begin{split}
  \textrm{Loss}(\mathcal{S}^*_i) = & -\smashoperator[r]{\sum_{\substack{(y,\mathcal{Q}^y)\in \mathbf{Train}\\ \textrm{s.t.} \,S^*_i \preceq [\mathcal{Q}^y]}}}\log(\textrm{Pr}[\mathcal{S}^*_i\,|\,y])\\
         & -\smashoperator[r]{\sum_{\substack{(y,\mathcal{Q}^y)\in \mathbf{Train}\\ \textrm{s.t.} \,\mathcal{S}^*_i \npreceq [\mathcal{Q}^y]}}}\log(1-\textrm{Pr}[\mathcal{S}^*_i\,|\,y]),
\end{split}
\end{equation}
where $\mathbf{Train}$ denotes the set of training data.

\subsection{Query Structure Ranking}
\label{sec:structureRank}
In this step, we use a combinational function to score each query structure in the training data for the input question. Since the prediction result for each query substructure is independent, the score for query structure $\mathcal{S}_i$ is measured by joint probability, which is
\begin{equation}
\begin{split}
  \textrm{Score}(\mathcal{S}_i\,|\,y) = &\smashoperator[r]{\prod_{\substack{\mathcal{S}^*_j\in\mathbf{FS^*}\\ \textrm{s.t.}\,  \mathcal{S}^*_j\preceq \mathcal{S}_i}}}{\textrm{Pr}[\mathcal{S}^*_j\,|\,y]} \\
  & \times  \smashoperator[r]{\prod_{\substack{\mathcal{S}^*_j\in\mathbf{FS^*}\\ \textrm{s.t.}\,  \mathcal{S}^*_j\npreceq \mathcal{S}_i}}}{(1-\textrm{Pr}[\mathcal{S}^*_j\,|\,y])}.
\end{split}
\end{equation}

Assume that $\mathcal{Q}^y\in \mathcal{S}_i$, $\forall \mathcal{S}^*_j\preceq \mathcal{S}_i$, we have $\mathcal{S}^*_j\preceq[\mathcal{Q}^y]$. Thus, $\textrm{Pr}[\mathcal{S}^*_j\,|\,y]$ should be 1 in the ideal condition. On the other hand, $\forall \mathcal{S}^*_j\npreceq \mathcal{S}_i$, $\textrm{Pr}[\mathcal{S}^*_j\,|\,y]$ should be 0. Thus, we have $\textrm{Score}(\mathcal{S}_i\,|\,y)=1$, and $\forall \mathcal{S}_k\neq \mathcal{S}_i$, we have $\textrm{Score}(\mathcal{S}_k\,|\,y)=0$.

\subsection{Query Substructure Merging}
\label{sec:merge}
We proposed a method, shown in Algorithm \ref{algo:merge}, to merge question-contained query substructures to build new query structures. In the initialization step, it selects some query substructures of high scores as candidates, since the query substructure may directly be the appropriate query structure for the input question. In each iteration, the method merges each question-contained substructures with existing candidates, and the merged results of high scores are used as candidates in the next iteration. The final output is the union of all the results from at most $K$ iterations.
\begin{algorithm}[!t]
\caption{Query substructure merging}
\label{algo:merge}
\SetCommentSty{textit}
{\small
\KwIn{Question $y$, freq. query substructures $\mathbf{FS}^*$}
	$\mathbf{FS}^+ := \{\mathcal{S}^*_i\in\mathbf{FS}^* \,|\, \textrm{Pr}[\mathcal{S}^*_i\,|\,y]>0.5\}$\;
	$\mathbf{M}^{(0)} := \{\mathcal{S}^*_i\in\mathbf{FS}^* \,|\, \textrm{Score}[\mathcal{S}^*_i\,|\,y]>\theta\}$\;
	\For(\tcp*[f]{$K$ is maximum iterations}){$i=1$ to $K$}{
		$\mathbf{M}^{(i)} := \emptyset$\;
		\ForAll{$\mathcal{S}^*_i\in\mathbf{FS}^+, \mathcal{S}_j\in\mathbf{M}^{(i-1)}$}{
			$\mathbf{M}^{(i)} := \mathbf{M}^{(i)} \cup \textrm{Merge}(\mathcal{S}^*_i,\mathcal{S}_j)$\;
		}
		$\mathbf{M}^{(i)} := \{\mathcal{S}_l\in\mathbf{M}^{(i)} \,|\, \textrm{Score}[\mathcal{S}_l\,|\,y]>\theta\}$\;
	}
	\Return $\bigcup_{i=0}^K \mathbf{M}^{(i)}$\;
}
\end{algorithm}

When merging different query substructures, we allow them to share some vertices of the same kind (variable, entity, etc.)  or edge labels, except the variables which represent aggregation results. Thus, the merged result of two query substructures is a set of query structures instead of one. Also, the following restrictions are used to filter the merged results:
\begin{enumerate}[label=(\roman*),itemsep=0pt]
\item The merged results should be connected;
\item The merged results have $\leq \tau$ triples;
\item The merged results have $\leq \delta$ aggregations;
\end{enumerate}

\begin{figure}[!t]
\centering
\includegraphics[width=0.92\columnwidth]{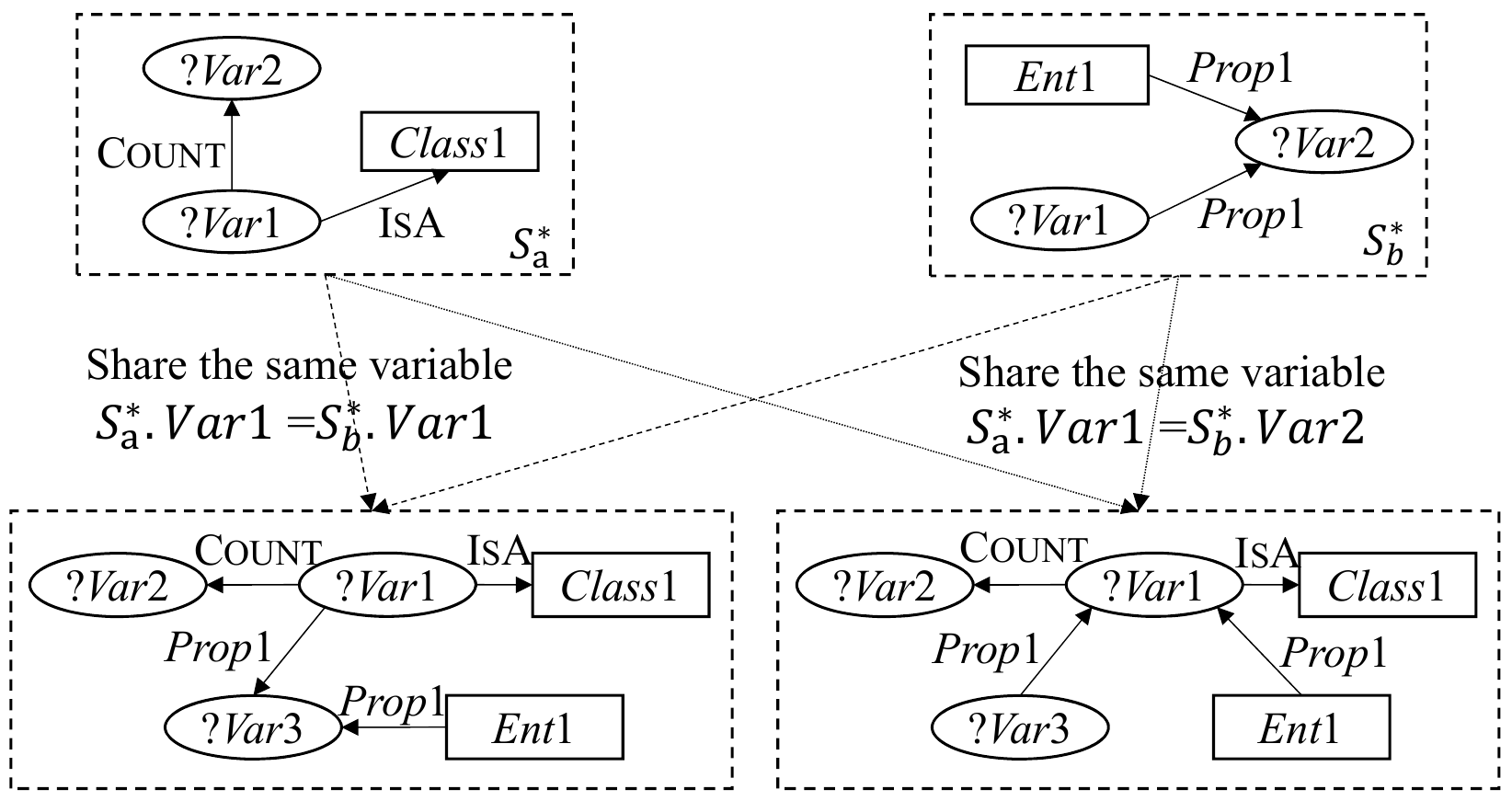}
\caption{Merge results for two query substructures}
\label{fig:merge}
\end{figure}

An example for merging two query substructures is shown in Figure \ref{fig:merge}.


\section{Experiments and Results}
In this section, we introduce the query generation datasets and state-of-the-art systems that we compare. We first show the end-to-end results of the query generation task, and then perform detailed analysis to show the effectiveness of each module. Question sets, source code and experimental results are available online.\footnote{\url{http://ws.nju.edu.cn/SubQG/}}
\subsection{Experimental Setup}
\paragraph{Datasets} We employed the same datasets as \citeauthor{QueryBuilding}(\citeyear{QueryBuilding}) and \citeauthor{SQG}(\citeyear{SQG}): (\romannumeral1) the large-scale complex question answering dataset (\textbf{LC-QuAD}) \cite{LC-QuAD}, containing 3,253 questions with non-empty results on DBpedia (2016-04), and (\romannumeral2) the fifth edition of question answering over linked data (\textbf{QALD-5}) dataset  \cite{QALD}, containing 311 questions with non-empty results on DBpedia (2015-10). Both datasets are widely used in KBQA studies \cite{gAnswer, EARL}, and have become benchmarks for some annual KBQA competitions\footnote{\url{http://lc-quad.sda.tech}}\footnote{\url{http://qald.aksw.org/index.php?q=5}}. We did not employ the WebQuestions \cite{SemanticParsing} dataset, since approximately 85\% of its questions are simple. Also, we did not employ the ComplexQuestions \cite{ConstraintQG} and ComplexWebQuestions \cite{DecompositionQuestion} dataset, since the existing works on these datasets have not reported the formal query generation result, and it is difficult to separate the formal query generation component from the end-to-end KBQA systems in these works.

%
%

\paragraph{Implementation details}
All the experiments were carried out on a machine with an Intel Xeon E3-1225 3.2GHz processor, 32 GB of RAM, and an NVIDIA GTX1080Ti GPU. For the embedding layer, we used random embedding. For each dataset, we performed 5-fold cross-validation with the train set (70\%), development set (10\%), and test set (20\%). The threshold $\gamma$ for frequent query substructures is set to $30$, the maximum iteration number $K$ for merging is set to $2$, $\theta$ in Algorithm~\ref{algo:merge} is set to $0.3$, the maximum triple number $\tau$ for merged results is set to $5$, and the maximum aggregation number $\delta$ is set to $2$.  Other detailed statistics are shown in Table \ref{tab:expSetup}.

\begin{table}[!t]
\captionof{table}{Datasets and implementation details}
\label{tab:expSetup}
\centering
\begin{tabular}{l|c|c}
  \hline & \small LC-QuAD & \small QALD-5 \\
  \hline \small No. of questions (complex) & \small  3,253 (2,249) & \small 311 (192) \\
   \small No. of query structures & \small 35 & \small 52 \\
   \small No. of freq. substructures & \small 37 & \small 10 \\
   \small Avg. training time & \small 1,102s & \small 272s  \\
   \small Avg. prediction time & \small 0.291s & \small 0.122s  \\
   \small Avg. query generation time & \small 0.356s & \small 0.197s  \\
  \hline
\end{tabular}
\end{table}
\subsection{End-to-End Results}
We compared SubQG with several existing approaches. SINA \cite{Sina} and NLIWOD conduct query generation by predefined rules and existing templates. SQG \cite{SQG} firstly generates candidate queries by finding valid walks containing all of entities and properties mentioned in questions, and then ranks them based on Tree-LSTM similarity. CompQA \cite{CQAEMNLP} is a KBQA system which achieved state-of-the-art performance on WebQuesions and ComplexQuestions over Freebase. We re-implemented its query generation component for DBpedia, which generates candidate queries by staged query generation, and ranks them using an encode-and-compare network.

The average F1-scores for the end-to-end query generation task are reported in Table \ref{tab:end2end}. All these results are based on the gold standard entity/relation linking result as input. Our approach SubQG outperformed all the comparative approaches on both datasets. Furthermore, as the results shown in Table \ref{tab:complex}, it gained a more significant improvement on complex questions compared with CompQA.
\begin{table}[t]
\centering
\caption{Average F1-scores of query generation}
\label{tab:end2end}
\begin{adjustbox}{width=\columnwidth}
\begin{tabular}{l|c|c}
	\hline & LC-QuAD & QALD-5 \\
	\hline Sina \cite{Sina} & $0.24\,^\dag$ & $0.39\,^\dag$ \\
   		  NLIWOD\,\footnotemark & $0.48\,^\dag$ & $0.49\,^\dag$ \\
		  SQG \cite{SQG} & $0.75\,^\dag$ & - \\
		  CompQA \cite{CQAEMNLP} & $0.772_{\pm 0.014}$ & $0.511_{\pm 0.043}$ \\
		  SubQG (our approach) & $\mathbf{0.846}_{\pm 0.016}$ & $\mathbf{0.624}_{\pm 0.030}$ \\
 	\hline
	\multicolumn{3}{c}{$^\dag$ indicates results taken from \citet{QueryBuilding} and SQG.} \\
\end{tabular}
\end{adjustbox}
\end{table}
\footnotetext{\url{https://github.com/dice-group/NLIWOD}}
\begin{table}[t]
\centering
\caption{Average F1-scores for complex questions}
\label{tab:complex}
\begin{adjustbox}{width=.7\columnwidth}
\begin{tabular}{l|c|c}
	\hline & LC-QuAD & QALD-5 \\
	\hline CompQA & $0.673_{\pm 0.009}$ & $0.260_{\pm 0.082}$ \\
		  SubQG & $\mathbf{0.779}_{\pm 0.017}$ & $\mathbf{0.392}_{\pm 0.156}$ \\
 	\hline
\end{tabular}
\end{adjustbox}
\end{table}

Both SINA and NLIWOD did not employ a query ranking mechanism, i.e., their accuracy and coverage are limited by the rules and templates. Although both CompQA and SQG have a strong ability of generating candidate queries, they perform not quite well in query ranking. According to our observation, the main reason is that these approaches tried to learn entire representations for questions with different query structures (from simple to complex) using a single network, thus, they may suffer from the lack of training data, especially for the questions with rarely appeared structures. As a contrast, our approach leveraged multiple networks to learn predictors for different query substructures, and ranked query structures using combinational function, which gained a better performance.

The results on QALD-5 dataset is not as high as the result on LC-QuAD. This is because QALD-5 contains 11\% of very difficult questions, requiring complex filtering conditions such as \textsc{Regex} and numerical comparison. These questions are currently beyond our approach's ability. Also, the size of training data is significant smaller.

\subsection{Detailed Analysis}
\subsubsection{Ablation Study}
\label{sec:comp}
We compared the following settings of SubQG:

\textbf{Rank w/o substructures.} We replaced the query substructure prediction and query structure ranking module, by choosing an existing query structure in the training data for the input question, using a BiLSTM multiple classification network.

\textbf{Rank w/ substructures} We removed the merging module described in Section \ref{sec:merge}. This setting assumes that the appropriate query structure for an input question exists in the training data.

\textbf{Merge query substructures} This setting ignored existing query structures in the training data, and only considered the merged results of query substructures.

As the results shown in Table \ref{tab:ablation}, the full version of SubQG achieved the best results on both datasets. \emph{Rank w/o substructures} gained a comparatively low performance, especially when there is inadequate training data (on QALD-5). Compared with \emph{Rank w/ substructures}, SubQG gained a further improvement, which indicates that the merging method successfully handled questions with unseen query structures.

Table \ref{tab:ablation2} shows the accuracy of some alternative networks for query substructure prediction (Section \ref{sec:subPredict}). By removing the attention mechanism (replaced by unweighted average), the accuracy declined approximately 3\%. Adding additional part of speech tag sequence of the input question gained no significant improvement. We also tried to replace the attention based BiLSTM with the network in \cite{Staged}, which encodes questions with a convolutional layer followed by a max pooling layer. This approach did not perform well since it cannot capture long-term dependencies.
\begin{table}[t]
\centering
\caption{Average F1-scores for different settings}
\label{tab:ablation}
\begin{adjustbox}{width=\columnwidth}
\begin{tabular}{l|c|c}
	\hline & LC-QuAD & QALD-5 \\
	\hline SubQG & $\mathbf{0.846}_{\pm 0.016}$ & $\mathbf{0.624}_{\pm 0.030}$ \\
	\hline Rank w/o substructures & $0.756_{\pm 0.012}$ & $0.383_{\pm 0.024}$ \\
	         Rank w/ substructures & $0.841_{\pm 0.014}$ & $0.614_{\pm 0.036}$ \\
	         Merge query substructures & $0.679_{\pm 0.020}$ & $0.454_{\pm 0.055}$ \\
 	\hline
\end{tabular}
\end{adjustbox}
\end{table}

\begin{table}[t]
\centering
\caption{Accuracy of query substructure prediction}
\label{tab:ablation2}
\begin{adjustbox}{width=\columnwidth}
\begin{tabular}{l|c|c}
	\hline & LC-QuAD & QALD-5 \\
	\hline BiLSTM w/ attention & $\mathbf{0.929}_{\pm 0.002}$ & $0.816_{\pm 0.010}$ \\
	\hline BiLSTM w/o attention & $0.898_{\pm 0.004}$ & $ 0.781_{\pm 0.009}$ \\
	         BiLSTM w/ attention + POS & $0.925_{\pm 0.004}$ & $\mathbf{0.818}_{\pm 0.007}$ \\
	         CNN in \cite{Staged} & $0.856_{\pm 0.006}$ & $0.740_{\pm 0.010}$ \\
 	\hline
\end{tabular}
\end{adjustbox}
\end{table}


\subsubsection{Results with Noisy Linking}
We simulated the real KBQA environment by considering noisy entity/relation linking results. We firstly mixed the correct linking result for each mention with the top-$5$ candidates generated from EARL \cite{EARL}, which is a joint entity/relation linking system with state-of-the-art performance on LC-QuAD. The result is shown in the second row of Table \ref{tab:noisy}. Although the precision for first output declined 11.4\%, in 85\% cases we still can generate correct answer in top-$5$. This is because SubQG ranked query structures first and considered linking results in the last step. Many error linking results can be filtered out by the empty query check or domain/range check.
\begin{table}[t]
\centering
\caption{Average Precision@$k$ scores of query generation on LC-QuAD with noisy linking}
\label{tab:noisy}
\begin{adjustbox}{width=\columnwidth}
\begin{tabular}{l|c|c}
	\hline & Precision@1 & Precision@5 \\
	\hline Gold standard & $0.842_{\pm 0.017}$ & $0.886_{\pm 0.014}$ \\
	         Top-5 EARL + gold standard & $0.728_{\pm 0.011}$ & $0.850_{\pm 0.009}$ \\
            Top-5 EARL  & $0.126_{\pm 0.012}$ & $0.146_{\pm 0.010}$ \\
 	\hline
\end{tabular}
\end{adjustbox}
\end{table}

We also test the performance of our approach only using the EARL linking results. The performance dropped dramatically in comparison to the first two rows. The main reason is that, for 82.8\% of the questions, EARL provided partially correct results. If we consider the remaining questions, our system again have 73.2\% and 84.8\% of correctly-generated queries in top-1 and top-5 output, respectively.

\subsubsection{Results on Varied Sizes of Training Data}
\begin{figure*}[htbp]
\centering
\includegraphics[width=\textwidth]{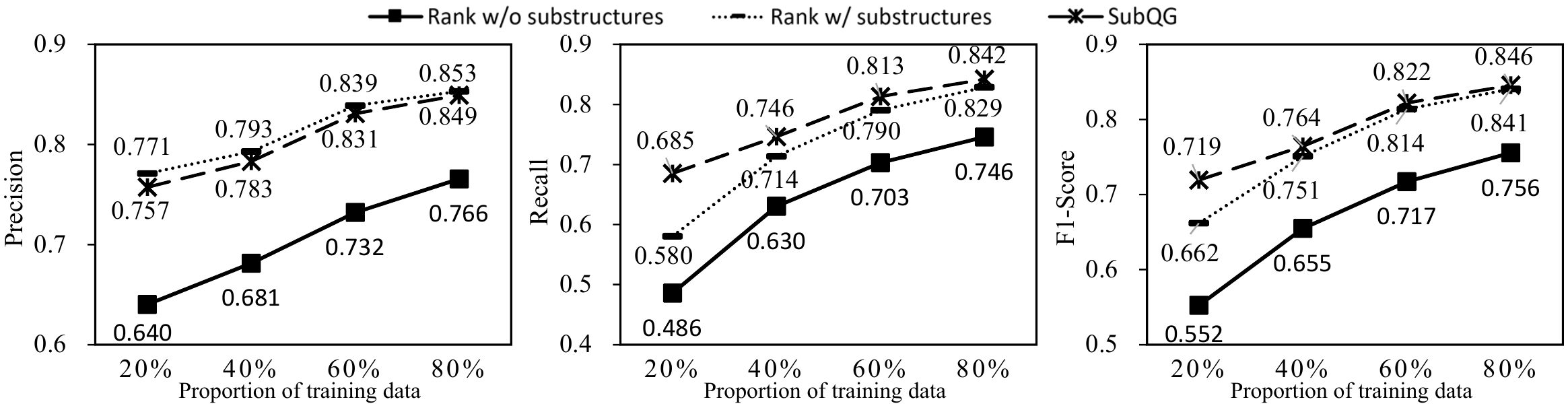}
\caption{Precision, recall and F1-score with varied proportions of training data}
\label{fig:trainingPercentage}
\end{figure*}

We tested the performance of SubQG with different sizes of training data. The results on LC-QuAD dataset are shown in Figure \ref{fig:trainingPercentage}. With more training data, our query substructure based approaches obtained stable improvements on both precision and recall. Although the merging module impaired the overall precision a little bit, it shows a bigger improvement on recall, especially when there is very few training data. Generally speaking, equipped with the merging module, our substructure based query generation approach showed the best performance.

\subsubsection{Error Analysis}
\label{sec:error}
We analyzed 100 randomly sampled questions that SubQG did not return correct answers. The major causes of errors are summarized as follows:

\textbf{Query structure errors (71\%)} occurred due to multiple reasons. Firstly, 21\% of error cases have entity mentions that are not correctly detected before query substructure prediction, which highly influenced the prediction result. Secondly, in 39\% of the cases a part of substructure predictors provided wrong prediction, which led to wrong structure ranking results. Finally, in the remaining 11\% of the cases the correct query structure did not appear in the training data, and they cannot be generated by merging substructures.

\textbf{Grounding errors (29\%)} occurred when SubQG generated wrong queries with correct query structures. For example, for the question \emph{``Was Kevin Rudd the prime minister of Julia Gillard"}, SubQG cannot distinguish $\langle JG, primeMinister, KR\rangle$ from $\langle KR,$ $primeMinister,JG\rangle$, since both triples exist in DBpedia. We believe that extra training data are required for fixing this problem.

\section{Related Work}
Alongside with entity and relation linking, existing KBQA systems often leverage formal query generation for complex question answering \cite{ConstraintQG,LC-QuAD}. Based on our investigation, the query generation approaches can be roughly divided into two kinds: \emph{template}-based and \emph{semantic parsing}-based.

Template-based approaches transform the input question into a formal query by employing pre-collected query templates. \citeauthor{KBQA}(\citeyear{KBQA}) collect different natural language expressions for the same query intention from question-answer pairs. \citeauthor{QueryBuilding}(\citeyear{QueryBuilding}) re-implement and evaluate the query generation module in NLIWOD, which selects an existing template by some simple features such as the number of entities and relations in the input question. Recently, several query decomposition methods are studied to enlarge the coverage of the templates. \citeauthor{TemplateQAWWW17}(\citeyear{TemplateQAWWW17}) present a KBQA system named QUINT, which collects query templates for specific dependency structures from question-answer pairs. Furthermore, it rewrites the dependency parsing results for questions with conjunctions, and then performs sub-question answering and answer stitching. \citeauthor{DecompositionTemplate}(\citeyear{DecompositionTemplate}) decompose questions by using a huge number of triple-level templates extracted by distant supervision. Compared with these approaches, our approach predicts all kinds of query substructures (usually 1 to 4 triples) contained in the question, making full use of the training data. Also, our merging method can handle questions with unseen query structures, having a larger coverage and a more stable performance.

Semantic parsing-based approaches translate questions into formal queries using bottom up parsing \cite{SemanticParsing} or staged query graph generation \cite{Staged}.
gAnswer \cite{gAnswer,gAnswer2} builds up semantic query graph for question analysis and utilize subgraph matching for disambiguation.
Recent studies combine parsing based approaches with neural networks, to enhance the ability for structure disambiguation. \citeauthor{ConstraintQG}(\citeyear{ConstraintQG}), \citeauthor{CQAEMNLP}(\citeyear{CQAEMNLP}) and \citeauthor{SQG}(\citeyear{SQG}) build query graphs by staged query generation, and follow an encode-and-compare framework to rank candidate queries with neural networks.
These approaches try to learn entire representations for questions with different query structures by using a single network. Thus, they may suffer from the lack of training data, especially for questions with rarely appeared structures. By contrast, our approach utilizes multiple networks to learn predictors for different query substructures, which can gain a stable performance with limited training data. Also, our approach does not require manually-written rules, and performs stably with noisy linking results.

\section{Conclusion}
In this paper, we introduced SubQG, a formal query generation approach based on frequent query substructures. SubQG firstly utilizes multiple neural networks to predict query substructures contained in the question, and then ranks existing query structures using a combinational function. Moreover, SubQG merges query substructures to build new query structures for questions without appropriate query structures in the training data. Our experiments showed that SubQG achieved superior results than the existing approaches, especially for complex questions.

In future work, we plan to add support for other complex questions whose queries require \textsc{Union}, \textsc{Group By}, or numerical comparison. Also, we are interested in mining natural language expressions for each query substructures, which may help current parsing approaches.

\section{Acknowledgments} 
This work is supported by the National Natural Science Foundation of China (Nos. 61772264 and 61872172). We would like to thank Yao Zhao for his help in preparing evaluation.

\bibliographystyle{acl_natbib}
\bibliography{emnlp2019}

\begin{thebibliography}{17}
\expandafter\ifx\csname natexlab\endcsname\relax\def\natexlab#1{#1}\fi

\bibitem[{Abujabal et~al.(2017)Abujabal, Yahya, Riedewald, and
  Weikum}]{TemplateQAWWW17}
Abdalghani Abujabal, Mohamed Yahya, Mirek Riedewald, and Gerhard Weikum. 2017.
\newblock \href {https://doi.org/10.1145/3038912.3052583} {Automated template
  generation for question answering over knowledge graphs}.
\newblock In \emph{Proceedings of the 26th International Conference on World
  Wide Web, {WWW} 2017}, pages 1191--1200.

\bibitem[{Bao et~al.(2016)Bao, Duan, Yan, Zhou, and Zhao}]{ConstraintQG}
Jun{-}Wei Bao, Nan Duan, Zhao Yan, Ming Zhou, and Tiejun Zhao. 2016.
\newblock \href {http://aclweb.org/anthology/C/C16/C16-1236.pdf}
  {Constraint-based question answering with knowledge graph}.
\newblock In \emph{Proceedings of the 26th International Conference on
  Computational Linguistics, {COLING} 2016}, pages 2503--2514.

\bibitem[{Berant et~al.(2013)Berant, Chou, Frostig, and
  Liang}]{SemanticParsing}
Jonathan Berant, Andrew Chou, Roy Frostig, and Percy Liang. 2013.
\newblock \href {http://aclweb.org/anthology/D/D13/D13-1160.pdf} {Semantic
  parsing on freebase from question-answer pairs}.
\newblock In \emph{Proceedings of the 2013 Conference on Empirical Methods in
  Natural Language Processing, {EMNLP} 2013}, pages 1533--1544.

\bibitem[{Cui et~al.(2017)Cui, Xiao, Wang, Song, Hwang, and Wang}]{KBQA}
Wanyun Cui, Yanghua Xiao, Haixun Wang, Yangqiu Song, Seung{-}won Hwang, and Wei
  Wang. 2017.
\newblock \href {https://doi.org/10.14778/3055540.3055549} {{KBQA:} learning
  question answering over {QA} corpora and knowledge bases}.
\newblock \emph{Proceedings of the VLDB Endowment}, 10(5):565--576.

\bibitem[{Dubey et~al.(2018)Dubey, Banerjee, Chaudhuri, and Lehmann}]{EARL}
Mohnish Dubey, Debayan Banerjee, Debanjan Chaudhuri, and Jens Lehmann. 2018.
\newblock \href {https://doi.org/10.1007/978-3-030-00671-6\_7} {{EARL:} joint
  entity and relation linking for question answering over knowledge graphs}.
\newblock In \emph{Proceedings of the 17th International Semantic Web
  Conference, {ISWC} 2018, Part {I}}, pages 108--126.

\bibitem[{Hu et~al.(2018)Hu, Zou, Yu, Wang, and Zhao}]{gAnswer2}
Sen Hu, Lei Zou, Jeffrey~Xu Yu, Haixun Wang, and Dongyan Zhao. 2018.
\newblock \href {https://doi.org/10.1109/TKDE.2017.2766634} {Answering natural
  language questions by subgraph matching over knowledge graphs}.
\newblock \emph{{IEEE} Transactions on Knowledge and Data Engineering},
  30(5):824--837.

\bibitem[{Luo et~al.(2018)Luo, Lin, Luo, and Zhu}]{CQAEMNLP}
Kangqi Luo, Fengli Lin, Xusheng Luo, and Kenny~Q. Zhu. 2018.
\newblock \href {https://aclanthology.info/papers/D18-1242/d18-1242} {Knowledge
  base question answering via encoding of complex query graphs}.
\newblock In \emph{Proceedings of the 2018 Conference on Empirical Methods in
  Natural Language Processing, {EMNLP} 2018}, pages 2185--2194.

\bibitem[{Raffel and Ellis(2015)}]{Attention}
Colin Raffel and Daniel P.~W. Ellis. 2015.
\newblock \href {http://arxiv.org/abs/1512.08756} {Feed-forward networks with
  attention can solve some long-term memory problems}.
\newblock \emph{CoRR}, abs/1512.08756.

\bibitem[{Shekarpour et~al.(2015)Shekarpour, Marx, Ngomo, and Auer}]{Sina}
Saeedeh Shekarpour, Edgard Marx, Axel{-}Cyrille~Ngonga Ngomo, and S{\"{o}}ren
  Auer. 2015.
\newblock \href {https://doi.org/10.1016/j.websem.2014.06.002} {{SINA:}
  semantic interpretation of user queries for question answering on interlinked
  data}.
\newblock \emph{Journal of Web Semantics}, 30:39--51.

\bibitem[{Singh et~al.(2018)Singh, Radhakrishna, Both, Shekarpour, Lytra,
  Usbeck, Vyas, Khikmatullaev, Punjani, Lange, Vidal, Lehmann, and
  Auer}]{QueryBuilding}
Kuldeep Singh, Arun~Sethupat Radhakrishna, Andreas Both, Saeedeh Shekarpour,
  Ioanna Lytra, Ricardo Usbeck, Akhilesh Vyas, Akmal Khikmatullaev, Dharmen
  Punjani, Christoph Lange, Maria{-}Esther Vidal, Jens Lehmann, and S{\"{o}}ren
  Auer. 2018.
\newblock \href {https://doi.org/10.1145/3178876.3186023} {Why reinvent the
  wheel: Let's build question answering systems together}.
\newblock In \emph{Proceedings of the 27th International Conference on World
  Wide Web, {TheWebConf} 2018}, pages 1247--1256.

\bibitem[{Talmor and Berant(2018)}]{DecompositionQuestion}
Alon Talmor and Jonathan Berant. 2018.
\newblock \href {https://aclanthology.info/papers/N18-1059/n18-1059} {The {Web}
  as a knowledge-base for answering complex questions}.
\newblock In \emph{Proceedings of the 2018 Conference of the North American
  Chapter of the Association for Computational Linguistics: Human Language
  Technologies, {NAACL-HLT} 2018}, pages 641--651.

\bibitem[{Trivedi et~al.(2017)Trivedi, Maheshwari, Dubey, and
  Lehmann}]{LC-QuAD}
Priyansh Trivedi, Gaurav Maheshwari, Mohnish Dubey, and Jens Lehmann. 2017.
\newblock \href {https://doi.org/10.1007/978-3-319-68204-4\_22} {{LC-QuAD}: {A}
  corpus for complex question answering over knowledge graphs}.
\newblock In \emph{Proceedings of the 16th International Semantic Web
  Conference, {ISWC} 2017, Part {II}}, pages 210--218.

\bibitem[{Unger et~al.(2015)Unger, Forascu, L{\'{o}}pez, Ngomo, Cabrio,
  Cimiano, and Walter}]{QALD}
Christina Unger, Corina Forascu, Vanessa L{\'{o}}pez, Axel{-}Cyrille~Ngonga
  Ngomo, Elena Cabrio, Philipp Cimiano, and Sebastian Walter. 2015.
\newblock \href {http://ceur-ws.org/Vol-1391/173-CR.pdf} {Question answering
  over linked data {(QALD-5)}}.
\newblock In \emph{Working Notes of Conference and Labs of the Evaluation
  Forum, {CLEF} 2015}.

\bibitem[{Yih et~al.(2015)Yih, Chang, He, and Gao}]{Staged}
Wen{-}tau Yih, Ming{-}Wei Chang, Xiaodong He, and Jianfeng Gao. 2015.
\newblock \href {http://aclweb.org/anthology/P/P15/P15-1128.pdf} {Semantic
  parsing via staged query graph generation: Question answering with knowledge
  base}.
\newblock In \emph{Proceedings of the 53rd Annual Meeting of the Association
  for Computational Linguistics, {ACL} 2015}, pages 1321--1331.

\bibitem[{Zafar et~al.(2018)Zafar, Napolitano, and Lehmann}]{SQG}
Hamid Zafar, Giulio Napolitano, and Jens Lehmann. 2018.
\newblock \href {https://doi.org/10.1007/978-3-319-93417-4\_46} {Formal query
  generation for question answering over knowledge bases}.
\newblock In \emph{Proceedings of the 15th Extended Semantic Web Conference,
  {ESWC} 2018}, pages 714--728.

\bibitem[{Zheng et~al.(2018)Zheng, Yu, Zou, and Cheng}]{DecompositionTemplate}
Weiguo Zheng, Jeffrey~Xu Yu, Lei Zou, and Hong Cheng. 2018.
\newblock \href {https://doi.org/10.14778/3236187.3236192} {Question answering
  over knowledge graphs: Question understanding via template decomposition}.
\newblock \emph{Proceedings of the VLDB Endowment}, 11(11):1373--1386.

\bibitem[{Zou et~al.(2014)Zou, Huang, Wang, Yu, He, and Zhao}]{gAnswer}
Lei Zou, Ruizhe Huang, Haixun Wang, Jeffrey~Xu Yu, Wenqiang He, and Dongyan
  Zhao. 2014.
\newblock \href {https://doi.org/10.1145/2588555.2610525} {Natural language
  question answering over {RDF:} a graph data driven approach}.
\newblock In \emph{Proceedings of the International Conference on Management of
  Data, {SIGMOD} 2014}, pages 313--324.

\end{thebibliography}

\end{document}